\documentclass[conference]{IEEEtran}
\IEEEoverridecommandlockouts
\usepackage{cite}
\usepackage{amsmath,amssymb,amsfonts}
\usepackage{algorithmic}
\usepackage{array,multirow,graphicx}
\newcommand{\STAB}[1]{\begin{tabular}{@{}c@{}}#1\end{tabular}}

\usepackage{textcomp}
\usepackage{xcolor}
\usepackage{color,soul}

\makeatletter
\def\ps@IEEEtitlepagestyle{%
    \def\@oddfoot{\mycopyrightnotice}%
    \def\@evenfoot{}%
}
\def\mycopyrightnotice{%
}

\let\old@ps@IEEEtitlepagestyle\ps@IEEEtitlepagestyle
\def\confheader#1{%
    \def\ps@IEEEtitlepagestyle{%
        \old@ps@IEEEtitlepagestyle%
        \def\@oddhead{\strut\hfill#1\hfill\strut}%
        \def\@evenhead{\strut\hfill#1\hfill\strut}%
    }%
    \ps@headings%
}
\makeatother

\confheader{%
        \parbox{20cm}{This paper is accepted and presented at\\ International Conference on Science and Contemporary Technologies (ICSCT) 2021.\\ It is to appear in the IEEE conference proceedings.}
}

\def\BibTeX{{\rm B\kern-.05em{\sc i\kern-.025em b}\kern-.08em
    T\kern-.1667em\lower.7ex\hbox{E}\kern-.125emX}}
\begin{document}


\title{Clustering Algorithms to Analyze the Road Traffic Crashes}

\author{\IEEEauthorblockN{Mahnaz Rafia Islam}
\IEEEauthorblockA{Dept. Computer Science and Engineering\\
University of Asia Pacific\\
Dhaka 1205, Bangladesh \\
17101007@uap-bd.edu}
\and
\IEEEauthorblockN{Israt Jahan Jenny}
\IEEEauthorblockA{Dept. Computer Science and Engineering\\
University of Asia Pacific\\
Dhaka 1205, Bangladesh \\
17101011@uap-bd.edu}
\and
\IEEEauthorblockN{Moniruzzaman Nayon}
\IEEEauthorblockA{Dept. Computer Science and Engineering\\
University of Asia Pacific\\
Dhaka 1205, Bangladesh \\
17101010@uap-bd.edu}
\and
\IEEEauthorblockN{Md. Rajibul Islam\thanks{Corresponding authors: md.rajibul.islam@uap-bd.edu and mamiruzzaman@wcupa.edu}}
\IEEEauthorblockA{Dept. Computer Science and Engineering\\
University of Asia Pacific\\
Dhaka 1205, Bangladesh \\
md.rajibul.islam@uap-bd.edu}
\and
\IEEEauthorblockN{Md Amiruzzaman}
\IEEEauthorblockA{Department of Computer Science\\
West Chester University\\
West Chester, PA 19383, USA\\
mamiruzzaman@wcupa.edu}
\and
\IEEEauthorblockN{M. Abdullah-Al-Wadud}
\IEEEauthorblockA{Department of Software Engineering\\
King Saud University\\
Riyadh 11543, Saudi Arabia\\
mwadud@ksu.edu.sa}
}

\maketitle


\begin{abstract}
Selecting an appropriate clustering method as well as an optimal number of clusters in road accident data is at times confusing and difficult. This paper analyzes shortcomings of different existing techniques applied to cluster accident-prone areas and recommends using Density-Based Spatial Clustering of Applications with Noise (DBSCAN) and Ordering Points To Identify the Clustering Structure (OPTICS) to overcome them. Comparative performance analysis based on real-life data on the recorded cases of road accidents in North Carolina also show more effectiveness and efficiency achieved by these algorithms.
\end{abstract}


\section{Introduction}
\begin{figure}[ht]
\centering
\includegraphics[width=0.75\columnwidth]{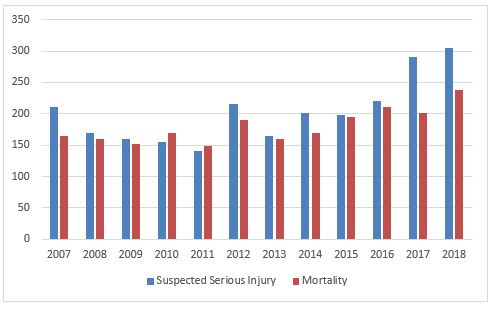}
\caption{Number of accidents}\label{fig:fig1}
\end{figure}

Traffic collision is the eighth preeminent reason of death worldwide, with an estimated 1.35 million people killed annually according to the  2018 Global Status Report on Road Safety  \cite{r1:who2018}. In the United States, road accident is a prominent cause of human death \cite{amiruzzaman2018prediction,r20:rizvee2021data}. In Fig. \ref{fig:fig1}, the major accident data from 2007 to 2018 in the North Carolina is visualized. The minimum number accidents causing suspected serious injuries is around 140, and that causing death is around 148 per occurring in a year during this period. 

A road traffic crash is an unpredictable event and can occur in various kinds of scenarios \cite{amiruzzaman2018prediction}. Types of crashes, environmental circumstances, highway configurations, vehicle features, and driver characteristics are among the many variables affecting road accidents. The main objective of accident data analysis is to recognize major parameters associated with road traffic accidents \cite{amiruzzaman2018prediction,r15:kumar2015data}. To provide safe driving instructions, road traffic statistics are critical to discovering variables related to fatal accidents. Data analysis can recognize various underlying reasons behind road accidents. The best solution is to study accident data in order to learn about the various causes of road accidents and to adopt the preventive steps. Again, the reasons as well as the types of accidents varies in different scenarios, making the analysis  a big challenge \cite{pasupathi2021trend,r20:rizvee2021data}. Therefore, to analyze the various circumstances of the occurrences of accidents, data mining methods like clustering algorithms, classification, and association rule mining, defining the different accident-prone geographical locations are very helpful in evaluating the different features relevant to road accidents \cite{r3:kumar2016data}. 

Clustering accident locations can be helpful to identify ac-cident hotspots \cite{r20:rizvee2021data,r19:anderson2009kernel}. Then, analyzing those hotspots might help administration to analyze them separately and deploy effective preventive mechanisms \cite{r22:prasannakumar2011spatio}. Also, if drivers know accident prone areas or hotspots, then they can also be more cautious in those areas to avoid accidents \cite{r21:lu2013clustering}. Awareness, cautiousness, safety, and prevention of traffic accident depend on accuracy of the clustering of traffic hotspots \cite{r19:anderson2009kernel,r20:rizvee2021data,r21:lu2013clustering}. Different algorithms for clustering, for example k-means clustering \cite{r16:puspitasari2020k}, fuzzy clustering \cite{r17:murat2012fuzzy} and hierarchical clustering \cite{r18:kumar2016analysis} are proposed to be used in clustering accident-prone locations. Most of such clustering algorithms suffer from the outliers, due to which the resultant clusters at times become sparse. However, for fruitful analysis, dense clusters are re-quired where different accident parameters are similar. 


In this paper, we look at the constraints of some widely used algorithms for clustering to classify high-frequency accident locations, and put forward DBSCAN and OPTICS algorithms as alternatives to overcome such limitations. Experimental evaluations based on a real life dataset \cite{r20:rizvee2021data} also show that these two algorithms outperform the other algorithms in clustering accident-prone places.

The following is how the remainder of the paper is organized: related work is described in section \ref{sec:relatedwork}, different clustering algorithms are discussed in section \ref{sec:clusteringalgorithm}. Section \ref{sec:methodology} explains different methods used to compare the performances of clustering algorithms. The experimental results are described in section \ref{sec:results}. Finally, section \ref{sec:conclusion} presents the conclusion.

\section{Related work}
\label{sec:relatedwork}
Clustering techniques have been used extensively by researchers and professional experts in the field of road accident for easy accident prediction, providing pedestrians with quick, adequate, accurate and less expensive safety delivery. Often studies reported that road accidents at certain geographic places are more common. For example, Kumar and Toshniwal \cite{r3:kumar2016data} used the K-means algorithm to identify high-frequency accident locations. They also indicated that before analyzing accident data, the use of appropriate clustering techniques reduces data variability and it can help to disclose hidden information. However, their experiment was limited to partitioning clustering algorithms only, as they ignored other algorithms for clustering. For instance, clustering algorithms based on density. In the partitioning clustering technique, a developer must define the number of clusters, which can lead to incorrect clustering of the given dataset. While clustering based on density selects the number of clusters by itself.

In \cite{r4:van2015using,r14:rendon2011internal} authors studied internal and external measurements to compare six types of clustering algorithms. The authors used (i) Silhouette measure, (ii) Davies-Bouldin measure, and (iii) Cali'nski-Harabasz measure methodology to compare clusters. This process can help a researcher to easily find which cluster is better for road accident data analysis. Moreover, the authors did not use the run time to compare clustering algorithms.  For non-expert users run time measurement can play an important role to choose an appropriate algorithm. We discuss all these important issues in this paper which can be very useful for a new researcher.

On different platforms (Python, Matlab, Wolfram and R) authors in \cite{r5:shahriar2019comparative} compared the performance of k-means and DBSCAN clustering algorithms. They used two criteria such as ``run time'' and ``accuracy'' to compare DBSCAN and k-means. However, at the end the authors did not show which clustering algorithm perform better in terms of ``run time'' and ``accuracy''. They just compared different algorithms on different platforms. But in our study, we discuss which algorithm is a better choice for road accident-based data analysis.

In a study, Patel and Thakral \cite{r6:patel2016best} compared some of clustering algorithms to find out the best clustering algorithm for data mining. At the end they also did not find a clustering algorithm which can produce the best result for specific data set. All the clustering algorithms performed well for different datasets. So their methodology is confusing for a new learner, in terms of choosing an appropriate algorithm for their dataset. In a seperate study, Shah and Jivani \cite{r7:shah2013comparison} also compared different clustering algorithm for data mining. They have used partitioning method, hierarchical method and density based method to compare clustering algorithms. At last they choose k-means algorithms for their research based on run time. But when it comes to accuracy then it is difficult to choose k-means for data mining.

\section{Clustering algorithms}
\label{sec:clusteringalgorithm}
\subsection{K-means} 
K-means is among the most commonly used clustering algorithms for compact clusters however it is vulnerable to outliers as well as noise, and it uses solely numerical attributes. It separates the dataset to $k$ clusters. The algorithm generates $k$ cluster centers at random, and each point is allocated to a cluster whose center is nearest to it. Cluster members are used to compute cluster centers. Since new centers vary from the previous ones, distances to cluster centers are computed again for each data point, and data points are allocated to the clusters with the shortest distance. The method of computing cluster centers and distances is repeated until the cluster centers do not shift substantially between calculations. To cluster, most K-Means type algorithms need the number of clusters ``k'' to be defined. One of the most significant drawbacks of these algorithms is this function.

\begin{figure}[h]
\centering
\includegraphics[width=.7\columnwidth]{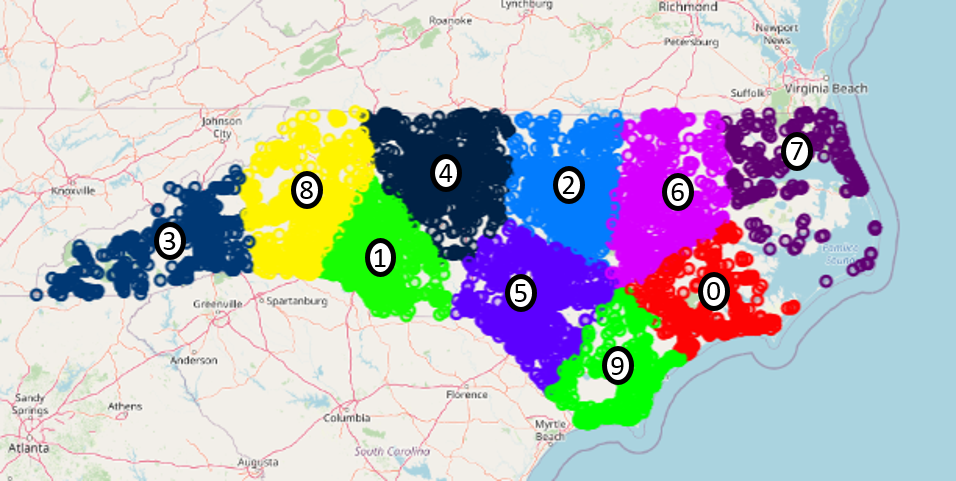}
\caption{Cluster visualization using k-means algorithm}\label{fig:fig2}
\end{figure}

There are some metrics to find an optimal number of $k$ such as the Elbow method and Silhoutte method. According to these two methods, we choose the optimal value of $k$ as 10 to get the best results from this algorithm. As shown in Fig. \ref{fig:fig2}, we visualized the clusters on a map in the form of a leaflet created by Python's ``Folium'' library.

Generally, the algorithms will often attempt to align an entity with the cluster's nearest center, they will build clusters of equal scale. Since the algorithm is designed to be more focused on cluster centers rather than cluster boundaries, this may also result in incorrect cluster creation. Another disadvantage of this algorithm is that it takes into account the entire dataset and is unable to differentiate between noise and cluster groupings. Due to its dependency on centroid placements, it can even classify clusters incorrectly.

\subsection{Mini-batch k-means}
\begin{figure}[h]
\centering
\includegraphics[width=.7\columnwidth]{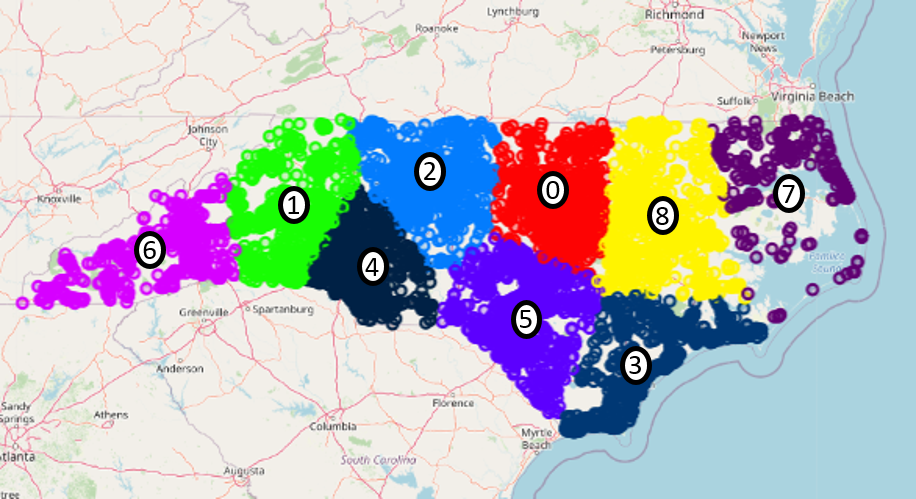}
\caption{Cluster visualization using Mini-batch k-means algorithm}\label{fig:fig3}
\end{figure}

The Mini Batch k-means \cite{r8:sculley2010web} was suggested as a better version of the k-means algorithm to cluster large datasets. The benefit of this version is that, the cost of processing is reduced by avoiding the use of the entire dataset per iteration. It uses a fixed-size subsample. Fig. \ref{fig:fig3} shows the clusters achieved from this algorithm with a batch size of 100.

The authors in \cite{r9:feizollah2014comparative} showed that in Android malware detection, this algorithm performs more effectively than the core version of k-means. We have used this algorithm as it is faster and memory-efficient.


\subsection{DBSCAN} 
Density-based clustering is also known as DBSCAN or Density-Based Spatial Clustering of Applications with Noise. It works by identifying a cluster as the most densely connected set of points possible \cite{r10:ester1996density}. Epsilon and minimum points are two criteria that must be taken into consideration. The maximal radius of the neighborhood is epsilon, and the minimal number of points in the epsilon-neighborhood to describe a cluster is minimum points. This clustering takes into account three different forms of points. Core, border, outlier are the three forms. 




There are some differences between Density-Based Clustering (DBSCAN) and K-means. For example, DBSCAN is excellent at eliminating noise from datasets by classifying them as Outliers, whereas K-means makes use of the entire dataset. K-means produces spherical shape clusters whereas DBSCAN gives arbitrarily shaped clusters. In a state, there can be both high accident-prone areas as well as low accident-prone areas. That is why clusters of arbitrary shapes can distinguish between them more efficiently.


\begin{figure}[h]
\centering
\includegraphics[width=.7\columnwidth]{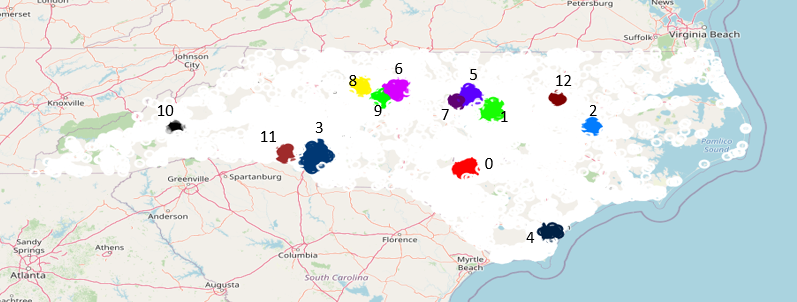}
\caption{Cluster visualization using DBSCAN algorithm}\label{fig:fig4}
\end{figure}

In our analysis, we clustered latitude-longitude data with scikit-learn's DBSCAN algorithm taking the eps value 5km, but it's divided by 6371 to convert it to radians as we have used the haversine distance metric which takes the distance in radians and the value of minimum sample was 300. It's worth mentioning that the eps and other factors were chosen at random. Fig. \ref{fig:fig4} demonstrates the cluster results.Provided latitudes and longitudes of two points, the haversine distance metric calculates the great-circle gap between those points on a region. We choose haversine because we want the distance to be determined over a 3D Earth. Moreover, latitude and longitude aren't distance units, but angles.

\subsection{OPTICS} 
The term ``optics'' refers to the use of ordering points to determine the clustering structure. The basic idea of the algorithm is to create a reachability chart. Each sample is assigned a distance of reachability and a point within the clusters ordering attributes. So these two attributes are only assigned when the model is adjust and are used to know which cluster it belongs to about the categorization it is based on density which means that it identifies different clusters within the data depending on how many points are clustered in a region. 

The most important features of the algorithm are: 

\begin{itemize}
    \item \textit{\textbf{Memory cost}}: optics algorithm requires more memory than other similar algorithms and this is because it keeps a priority queue to know which is the next data points and 
    
    \item \textit{\textbf{Few parameters}}: It uses very few parameters because it does not require to maintain the epsilon that in addition is only given in the pseudo-code to reduce the time taken. 
    
\end{itemize}

\begin{figure}[h]
\centering
\includegraphics[width=0.7\columnwidth]{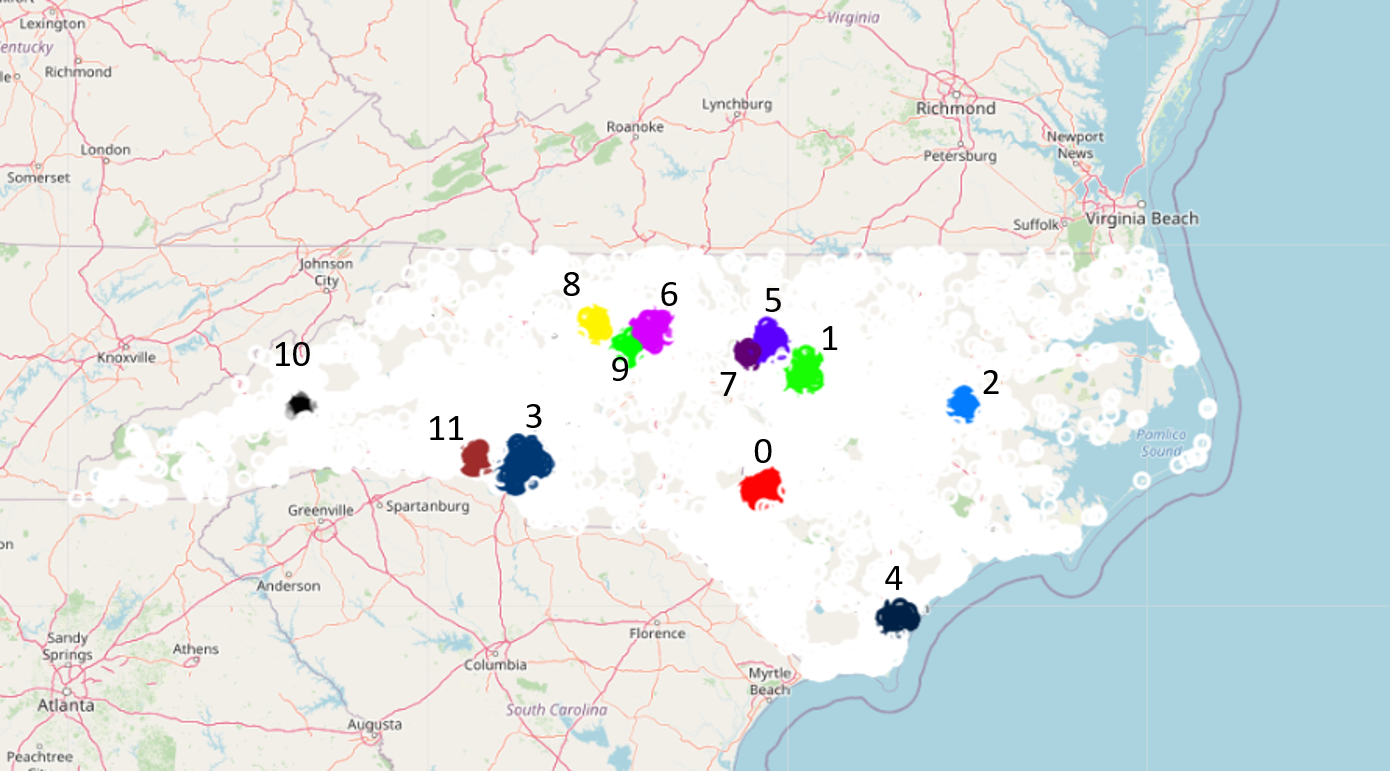}
\caption{Cluster visualization using OPTICS algorithm}\label{fig:fig5}
\end{figure}

Optics only works with numerical data. one major advantage of this approach is that compared to other clustering algorithms it does not limit itself to just one global parameter setting instead documented cluster ordering contains information which is equivalent to the density-based clustering corresponding to a broad range of parameter settings and thus is a flexible foundation for clustering, both automated and interactive. Fig. \ref{fig:fig5} shows the outlook of the result obtained from this algorithm. One of the disadvantages of this algorithm is that it is not suitable for high-dimensional spaces. 


\section{METHODOLOGY}
\label{sec:methodology}
\begin{figure}[h]
\centering
\includegraphics[width=0.7\columnwidth]{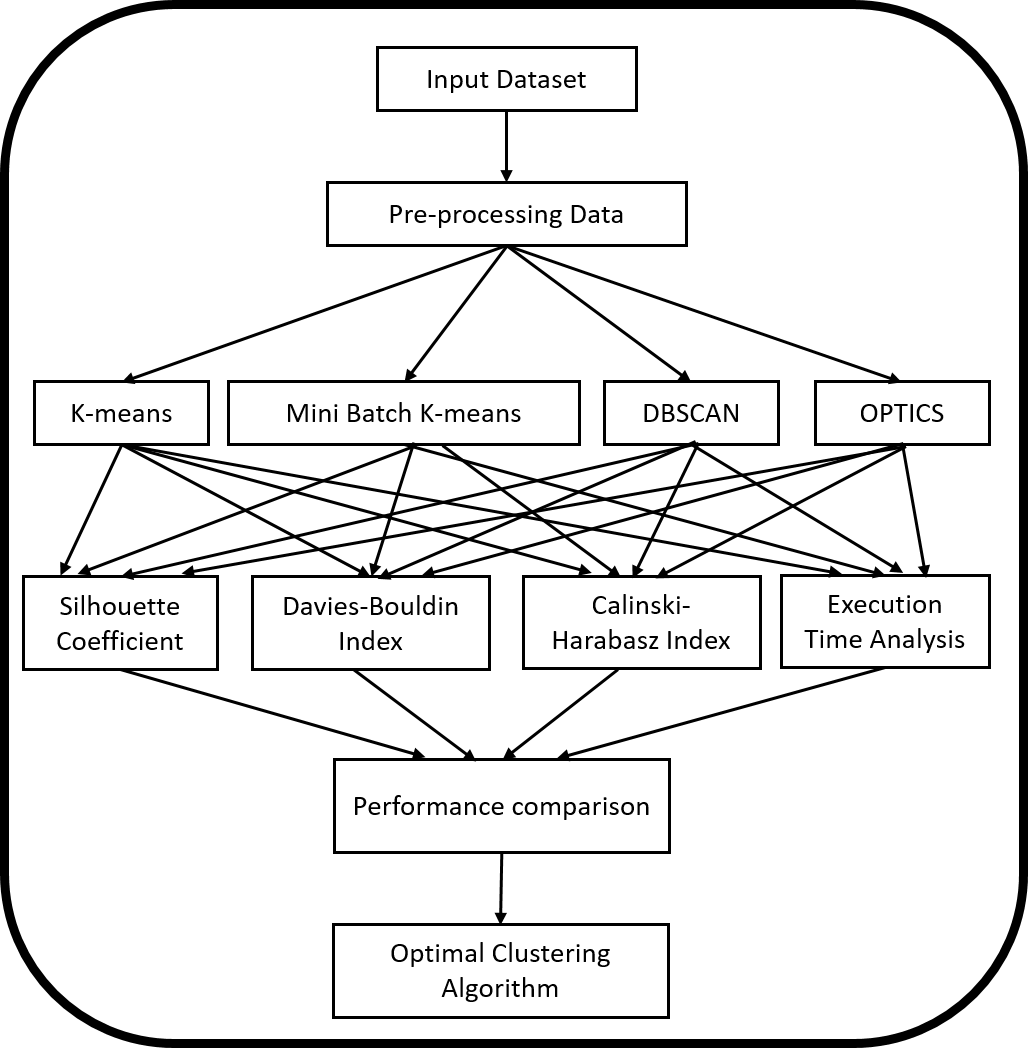}
\caption{Step-by-step flow of this research study}\label{fig:fig6}
\end{figure}

Estimating how well an algorithm for clustering has performed is not so easy as counting the amount of errors, accuracy, and recall, as it is with supervised learning algorithms. In this case, clusters are assessed using a metric of similarity or dissimilarity, such as the distance between two cluster points. The distance of data points between two different clusters is maximized, while the distance between the same clusters is minimized. The algorithm has done well if it can break off dissimilar observations and group similar observations together. Fig. \ref{fig:fig6} shows the overall workflow of our analysis. 

We begin by entering the dataset into our system. The data is then pre-processed by deleting any extraneous columns. We simply need the latitude and longitude values for this analysis. Two types of performance assessment are used for clustering techniques. One is external assessment in which we have labels in the data set and another is internal assessment in which there is no label in the dataset. As our dataset belongs to the category of no labels, so we have used some validity indices that are defined in section IV for internal assessment. We have used three most commonly used machine learning model evaluation metrics to effectively assess the performance of the clustering algorithms.

\subsection{Silhouette Coefficient} 
The silhouette Coefficient \cite{r11:rousseeuw1987silhouettes} is a metric used to calculate the goodness of a clustering technique. It demonstrates how close an object is to the other objects in its cluster (cohesion) also how distinct or well separated it is from a different clusters (separation). The silhouette score is confined from $-1$ to $+1$, where a higher value means the object is well matched in its cluster as well as a bad match to its neighboring clusters. Whereas, a lower value means that the object is a bad match in its cluster as well as have some similarity to its neighboring clusters. As shown in Fig. \ref{fig:fig8}, we get the highest value of 0.751 from DBSCAN. 

\begin{figure}[h]
\centering
\includegraphics[width=0.70\columnwidth]{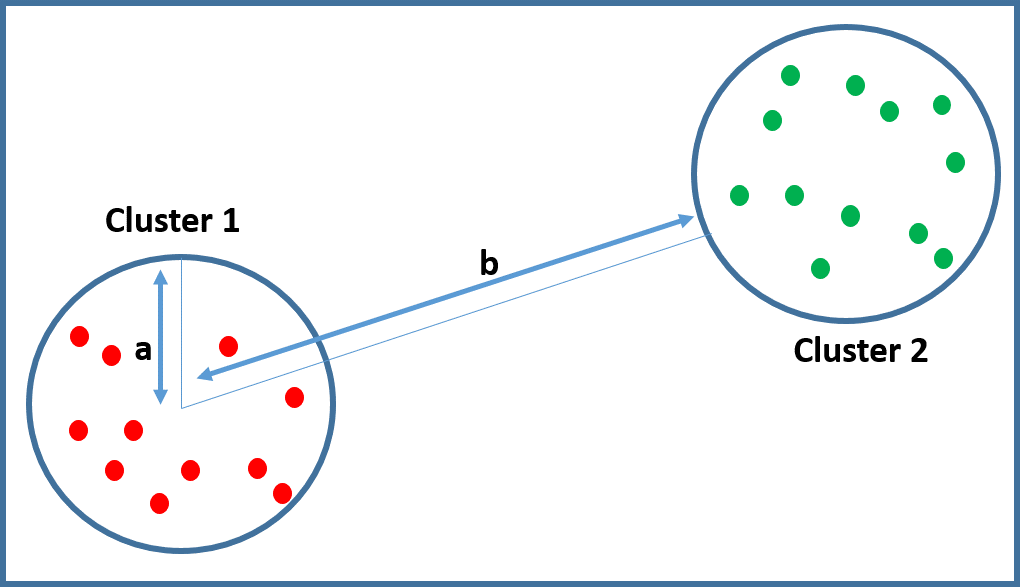}
\caption{Showing computational process of Silhouette parameters: $a$ within cluster distance and $b$ between cluster distance. General expectation is $a$ should be small and $b$ should be large}\label{fig:fig7}
\end{figure}

The silhouette score is composed of two distances:

\begin{equation}
    s = \frac{b-a}{\max{(a,b)}}
\end{equation}

In Fig. \ref{fig:fig7}, a is intra-cluster distance which is the mean distance between each point within a cluster, and b is the inter-cluster distance which is the mean distance between a sample and the next closest cluster's other points. In the ideal case, $a \leq b$ is expected for a good clustering.


\subsection{Davies-Bouldin Index} 
This index \cite{r12:davies1979cluster} is dependent on a ratio between “within-cluster” and “between-cluster” distances:
\begin{equation}
    V_{DB} = \frac{1}{2}\sum_{i=1}^{k} \max_{j\leq k, j\neq i} D_{ij}
\end{equation}
Let's assume we have five clusters 0, 1, 2, 3 and 4. We calculate $D_{ij}$ between cluster 0 and 1, 0 and 2, 0 and 3, 0 and 4 and we take the maximum of those values. We do the same for cluster 1, 2, 3 and 4. We then calculate the average of those maximum values. $D_{ij}$  is the "within-to-between cluster distance ratio" for the i and j clusters.

\begin{equation}
    D_{ij} = \frac{(\bar{d_i}+\bar{d_j})}{d_{ij}}
\end{equation}

Here, $d_i$ is the mean distance between every single data points in cluster $i$ and its centroid, similar for $d_j$. $d_{ij}$ is the distance between the centroids of the two clusters. Thus $D_{ij}$ is the cluster similarity index here and that is defined as a result of the standard deviations being added together or the dispersions divided by the difference of the center vectors. 

Now if the sum of $d_i$ and $d_j$ is small and $d_{ij}$ is large this indicates a small value of $i$ and $j$ cluster similarity. According to Davies-Bouldin Index, one means the two clusters touch each other, a value larger than one means there is an overlap between the clusters and a value smaller than one indicates nice separation of the clusters. So now this is a measure for how well two clusters are separated and we want small values obviously. As shown in Fig. \ref{fig:fig9}, we get the smallest value of 0.256 from DBSCAN.

Intra-cluster similarity and inter-cluster differences are calculated by the Davies-Bouldin Index, while the Silhouette score calculates the distance between every single data points, the cluster's center point to which it was allocated and the nearest centroid belonging to another cluster.

\subsection{Calinski-Harabasz Index} 

This index \cite{r13:calinski1974dendrite} also known as the Variance Ratio Criterion, is the ratio of the sum of ``inter-clusters dispersion'' and ``intra-cluster dispersion'' for every clusters. There is no specific range for the score of this index, the ``higher'' the score is, the ``better'' the algorithm has performed. A higher score indicates that the clusters are dense as well as distinct from one another, which is linked to a generic cluster definition. As shown in Fig. \ref{fig:fig10}, we get the highest value of 597,722.21 from OPTICS.

Let's assume we have $k$ clusters, each with its own relative centroids, as well as the global centroid. The inter-cluster dispersion, $SS_B$ and intra-cluster dispersion, $SS_w$ is refereed to as:

\begin{equation}
    SS_w = \sum_{i}^{k}\sum_{x \in c_i} ||x - m_i||^2
\end{equation}

\begin{equation}
    SS_B = TSS - SS_w
\end{equation}

Here, $k$ represents how many clusters are there, $N$ signifies the total number of data points and $TSS$ is the total sum of squares. The total number of squares for a given dataset is the square distance of all the data points from the centroid of the dataset. Also, $x$ is each data point, $C_i$  means $i$th cluster, $m_i$ is the centroid of cluster $i$, and $||x-m_i ||$ is the distance between the two vectors. Thus the Calinski-Harbasz Index (CHI) is followed up by the ratio of the between cluster dispersion and the within-cluster dispersion:

\begin{equation}
    CHI = \frac{SS_B}{SS_W} \times \frac{N-k}{k-1}
\end{equation}

Here, $N$ is total number of data points.
The Calinski-Harabasz index tests the validity of the cluster using the average sum of squares inside and between clusters, while the Silhouette index validates the efficiency of the cluster using the pair by pair difference within and between cluster distances.

\subsection{Execution time} 
There is another important part which will be about execution time. When we are completing any analysis on small dataset it will be cost effective but for any huge dataset we have to check the cost effectiveness through power consumption and by calculating execution time according to several clustering algorithms to identify that in which clustering algorithm it takes less run time. So we have calculated run time based on our dataset by implementing four types of clustering algorithms and the results are shown in Table \ref{table1}.

\section{Results}
\label{sec:results}
The results of clustering algorithms that we have chosen for this comparison is shown in Table \ref{table1}.

\begin{table}[htbp]
\caption{Comparisons of clustering methods: results are shown smallest to largest}\label{table1}
\resizebox{\columnwidth}{!}{\begin{tabular}{l|l|l|l|l}
\hline
                                                                     & \parbox[t]{2cm}{K-means\\ (10 clusters)} & \parbox[t]{2cm}{Mini-Batch\\ K-means\\ (9 clusters)} & \parbox[t]{2cm}{DBSCAN\\(13 clusters)} & \parbox[t]{2cm}{OPTICS\\(12 clusters)} \\ \hline
\multirow{13}{*}{\STAB{\rotatebox[origin=c]{270}{ Largest to smallest cluster number}}} & 8519   (1)           & 8521   (4)                    & 5435   (3)          & 5435   (3)          \\ 
                                                                     & 6759   (2)           & 6816   (0)                    & 2956   (1)          & 2956   (1)          \\ 
                                                                     & 5437   (4)           & 5442   (2)                    & 1870   (6)          & 1870   (6)          \\ 
                                                                     & 3218   (5)           & 3340   (5)                    & 1664   (5)          & 1664   (5)          \\ 
                                                                     & 2332   (6)           & 2796   (8)                    & 1327   (0)          & 1327   (0)          \\ 
                                                                     & 2079   (8)           & 2364   (3)                    & 948    (4)          & 948   (4)           \\ 
                                                                     & 1913   (3)           & 2079   (1)                    & 829   (10)          & 827   (10)          \\ 
                                                                     & 1660    (9)          & 1913   (6)                    & 823     (8)         & 821   (8)           \\ 
                                                                     & 1333    (0)          & 436    (7)                    & 580    (11)         & 579   (11)          \\ 
                                                                     & 457     (7)          &                               & 563    (9)          & 562    (9)          \\ 
                                                                     &                      &                               & 467    (2)          & 467    (2)          \\ 
                                                                     &                      &                               & 357    (7)          & 357    (7)          \\ 
                                                                     &                      &                               & 321   (12)          &                     \\ \hline
\end{tabular}
}
\end{table}

Now, if we look at the highest accident prone areas that have clustered together using different clustering methods, it is  visible that there is some difference at the number of data points in each largest clusters which can confuse the audience to understand highly accident-prone areas. That is why we have performed some comparisons in this paper to find out an optimum clustering algorithm for road traffic crashes data.

In this segment, we calculate the efficiency of these algorithms using two types of criteria: internal cluster validation metric and execution time.

\subsection{Internal Evaluation}

\begin{figure}[h]
\centering
\includegraphics[width=0.65\columnwidth]{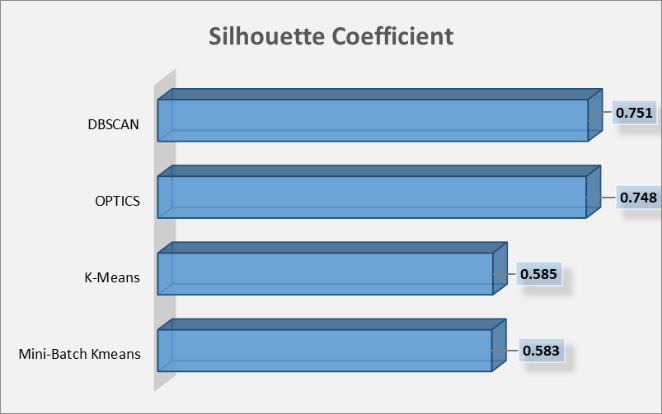}
\caption{Comparison of clustering methods in terms of Silhouette Coefficient}\label{fig:fig8}
\end{figure}


\begin{figure}[h]
\centering
\includegraphics[width=0.65\columnwidth]{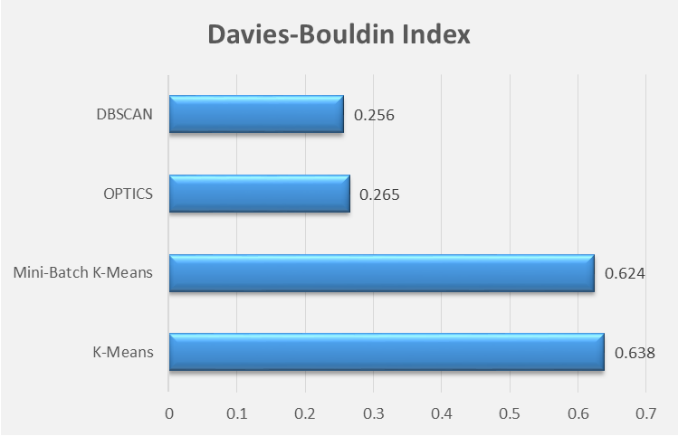}
\caption{Comparison of clustering methods in terms of Davies-Bouldin Index}\label{fig:fig9}
\end{figure}

\begin{figure}[h]
\centering
\includegraphics[width=0.65\columnwidth]{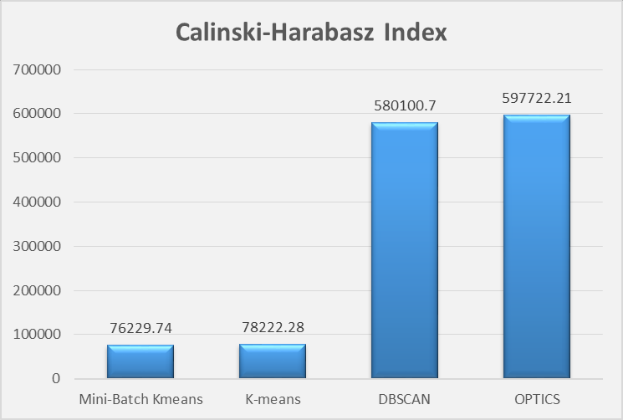}
\caption{Comparison of clustering methods in terms of Calinski-Harabasz Index}\label{fig:fig10}
\end{figure}

From the observations, we can say that DBSCAN is a better choice according to the Silhouette Metric and Davies-Bouldin Index. For convex clusters, the Calinski-Harabasz index is typically higher than for other cluster definitions, such as clusters based on density, like clusters gleaned from DBSCAN \cite{r20:rizvee2021data}. Closest mean is used to segment the space in K-means: the clusters create convex areas. A group of points on a plane is considered convex if a line segment connecting some two points lies inside the set. In DBSCAN and OPTICS, we define a maximum radius with which to form clusters. The algorithm will scan the space and group together points that are all reachable from one another. However, we can sometimes end up with a non-convex cluster. So now if we compare the Calinski-Harabasz Index of Optics and DBSCAN, we can see that OPTICS has a little higher score than DBSCAN, which we have considered as negligible.

\subsection{Execution time analysis} 

The execution time of these algorithms is implemented using an in-built python library named time it on a \textbf{core i3 processor} with 1.90 GHz clock speed and 8GB of RAM.



\begin{table}[h!]
\caption{CLUSTERING EXECUTION TIME FOR DIFFERENT ALGORITHMS}\label{table2}
\resizebox{\columnwidth}{!}{\begin{tabular}{l|l|l|l|l}
\hline
\multirow{2}{*}{\parbox[t]{2cm}{Total no of\\ test cases}} & \multicolumn{4}{l}{Time taken to cluster (Sec)} \\ \cline{2-5} 
                                      & k-means  & \parbox[t]{2cm}{Mini-Batch\\ k-means}  & OPTICS  & DBSCAN  \\ \hline
10000                                 & 0.202    & 0.147             & 13.349  & 1.626   \\ 
21854                                 & 0.301    & 0.15              & 43.675  & 5.091   \\ 
33707                                 & 0.398    & 0.389             & 83.656  & 8.399   \\ \hline
\end{tabular}
}
\end{table}

It is now clearly visible that DBSCAN takes much less time than OPTICS to create clusters (see Table \ref{table2}).

\section{CONCLUSION}
\label{sec:conclusion}
In this paper, focusing on overcoming the constraints of different existing proposals for clustering accident-prone areas, we suggest  DBSCAN and OPTICS. Since both DBSCAN and OPTICS are density-based clustering algorithms,  they can help to find spatial clusters and handle outliers better than the other algorithms used in this domain.  Experimental evaluation on a real life dataset also advocates for the same.

\section*{Acknowledgment}
We would like to thank the Institute of Energy, Environment, Research, and Development (IEERD, UAP) and the University of Asia Pacific for financial support.

\bibliographystyle{IEEEtran}      
\bibliography{reference}

\end{document}